# Learning to Lead Themselves: Agentic AI in MAS using MARL

Ansh Kamthan [*][‡]

[*]Manipal University Jaipur, Department of Artificial Intelligence and Machine Learning, Rajasthan

[‡]Corresponding author: ansh.229310229@muj.manipal.edu

## ABSTRACT

As autonomous systems move from prototypes to real deployments, the ability of multiple agents to make decentralized, cooperative decisions becomes a core requirement. This paper examines how agentic artificial intelligence, agents that act independently, adaptively and proactively can improve task allocation and coordination in multi-agent systems (MAS), with primary emphasis on drone delivery and secondary relevance to warehouse automation. We formulate the problem in a cooperative multi-agent reinforcement learning (MARL) setting and implement a lightweight multi-agent Proximal Policy Optimization (called IPPO) approach in PyTorch under a centralized-training, decentralized-execution paradigm. Experiments are conducted in PettingZoo's simple_spread_v3 environment, where multiple homogeneous "drones" or "agents" must self-organize to cover distinct targets without explicit communication.

Across training, agents learn decentralized policies that exhibit improvements in team reward and emergent spatial separation, indicative of effective task allocation. We provide quantitative and qualitative evidence of coordination as training curves, behaviour visualizations, analyses of reward stability, policy characteristics and discuss design trade-offs that influence convergence and robustness. Finally, we connect these results to real-world constraints in drone fleets and warehouse robotics for deploying agentic systems. Overall, this work offers an early, implementable step toward scalable, self-managing multi-agent coordination, highlighting both the promise and the open challenges of agentic AI in cooperative environments.

# 1. INTRODUCTION

Many real-world systems like autonomous drones[1], warehouse robots[2] and decentralized delivery fleets[3], require agents that can perceive their surroundings, reason about tasks and coordinate to achieve shared objectives. As autonomy[4] and adaptability improve, the role of AI agents in dynamic multi-agent settings has grown in both technical and strategic importance.

We use the term agentic systems for agents that make independent decisions, respond to new information and adjust strategies over time. Unlike models that produce static outputs, these agents act sequentially and interact with other agents and the environment[5]. In this paper, we focus on agentic AI[6] that exhibits autonomy, proactivity and decentralized coordination in cooperative tasks such as allocating work among delivery drones.

A natural path to such behaviour is Multi-Agent Reinforcement Learning (MARL)[7], [8], where multiple learners share an environment and must cope with non-stationarity, decentralized policies[9] and credit assignment in team rewards. Prior approaches include value-based methods such as QMIX[10], actor-critic methods with centralized critics such as MADDPG[11] and scalable policy-gradient methods such as Proximal Policy Optimization (PPO)[12], [13].

This paper investigates how decentralized coordination can emerge among drone agents trained with multi-agent PPO, which we called IPPO, by implementing a lightweight, custom PyTorch[14] setup that avoids heavyweight frameworks to keep the pipeline transparent and controllable. Our experiments use PettingZoo's simple_spread_v3[15], a cooperative task in which multiple drones must distribute themselves in space and allocate coverage dynamically. The goal is not only to train effective policies but also to characterize the patterns of emergent coordination, agentic autonomy and decentralized decision-making observed during training.

Our study is intentionally scoped to a stylized cooperative coverage task with homogeneous agents and modest training budgets. Results should be interpreted as a lightweight, reproducible baseline and as design guidance for coordination in cooperative MARL, rather than as a state-of-the-art claim.

This paper makes the following contributions aligned with the current content:

1. A compact multi-agent PPO baseline for cooperative coverage in simple_spread_v3, implemented in PyTorch without reliance on heavy frameworks.

2. Descriptive evidence of emergent coordination, rising team reward and spatial separation, based on the included plots and behaviour visualizations.

3. A clear training protocol and diagnostics (hyperparameters and learning curves) to support replication on standard hardware.

4. A deployment-oriented discussion mapping observed behaviours to drone delivery and warehouse contexts.

## 2. Related Work & Background

### 2.1 Agentic AI & Autonomous Agents

The pursuit of autonomous intelligent systems capable of reasoning, adapting and acting across time horizons has been key to the field of artificial intelligence since its origin. In recent years, a surge of interest has emerged around agentic AI[6], systems with high degrees of autonomy, often embedded in dynamic, open-ended environments where they must act independently to achieve complex goals[5].

Agentic systems differ from traditional AI models (Reactive Systems)[16] in that they are not merely predictive or generative tools, such as image classifiers or text summarizers but actors, entities capable of initiating actions, learning from consequences and engaging in ongoing interaction loops with their environments. Such systems may include reinforcement learning agents, robotics controllers or language model agents with tool access and persistent state memory.

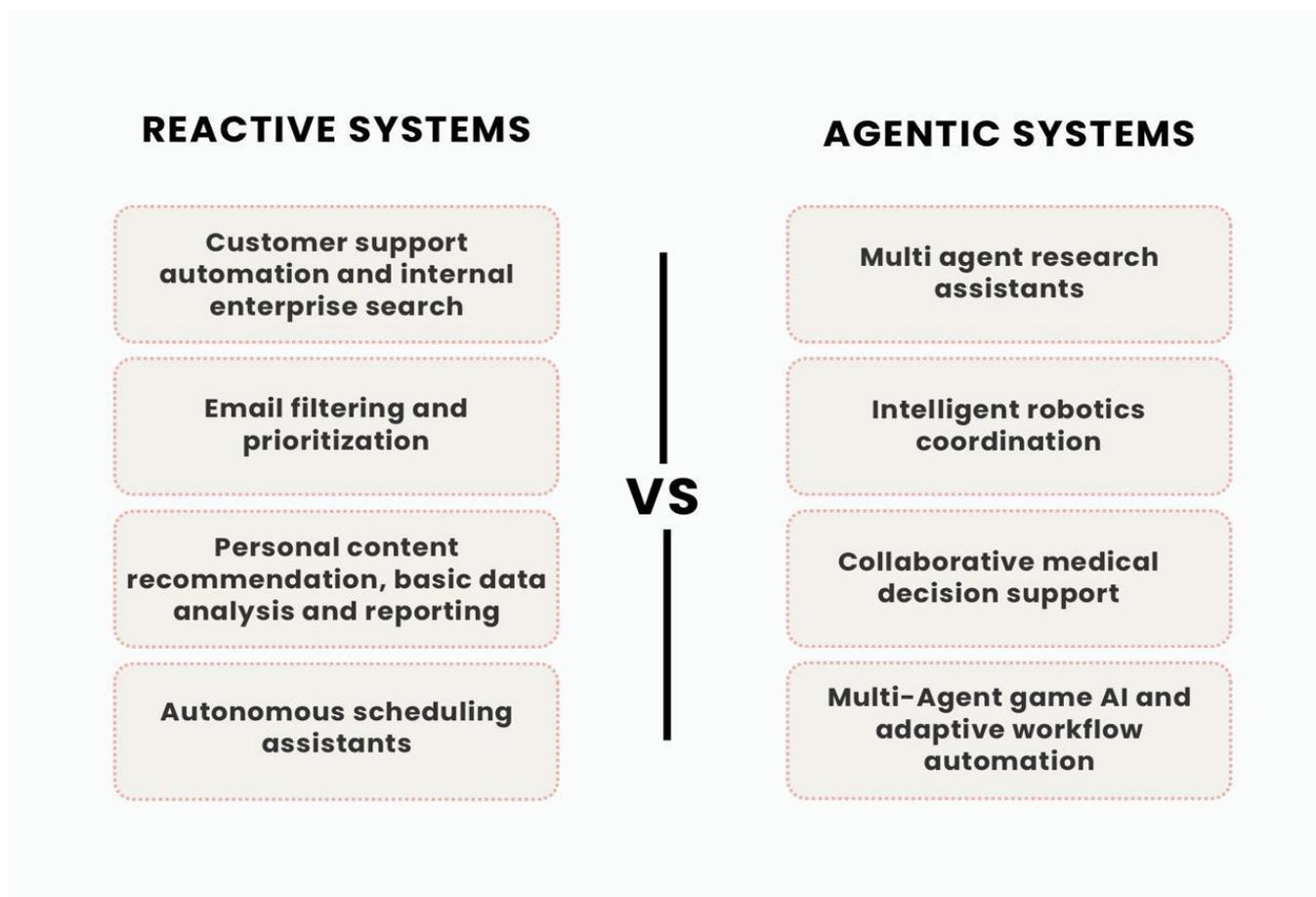

Figure 1 : Reactive Systems Vs Agentic Systems

This shift toward agentic capabilities is driven by multiple factors, including improvements in policy-optimization methods, scaling laws in large models, increasing economic and industrial demand for autonomous decision-making and the desire to develop systems that can reliably operate under limited supervision. From household assistants to scientific experimenters and autonomous traders, the scope of agent-based deployment is rapidly growing.

While the potential of agentic systems is profound, it does come with risks and technical challenges. Without careful design, agentic systems can exhibit misalignment or unexpected behaviours, especially when operating in multi-agent contexts. Hence, visibility into agent behaviour, strategies and training dynamics is not only a technical requirement but a governance imperative.

Understanding how agentic capabilities emerge, through feedback[8], [17], adaptation and interaction, remains a central concern in artificial intelligence research. The increasing deployment of agentic systems in several important domains demands robust frameworks for evaluating autonomy. Once these systems operate without a human in the loop, their ability to reason and learn becomes the ballgame. Studying these behaviours in structured multi-agent simulations offers a rich insight into the layered dynamics of agency, where policy, perception and decision-making converge under shared limits.

**Table 1:** *Reactive Systems vs. Agentic Systems*

| Characteristic | Reactive Systems | Agentic Systems |
| --- | --- | --- |
| Decision Basis | Immediate input | Long-term goals + internal state |
| Autonomy | Low | High |
| Adaptability | Limited (pre-programmed) | Learns from environment |
| Task Horizon | Single step | Sequential/multi-step |
| Environment Interaction | One-time or static | Ongoing and dynamic |
| Examples | Classifiers, image generators | RL agents, multi-agent systems, tool-using LLMs |

[16]

**2.2 Multi-Agent Reinforcement Learning (MARL)**

MARL extends the standard reinforcement learning model to systems involving multiple interacting agents. Each agent must learn a policy that not only adapts to its own environment but also to the behaviours of other agents, which themselves are learning simultaneously. This introduces non-stationarity into the environment, ravelling credit assignment, stability and convergence[7], [8], [17]

MARL algorithms can be broadly categorized by how they approach agent coordination and training:

- **Independent Learners** treat each agent as an isolated learner[18], using standard RL techniques.

- **Centralized Training with Decentralized Execution (CTDE)**[11], [19] is a popular paradigm where agents are trained using a shared global state or centralized critic but execute policies based on local observations.

Popular MARL algorithms include:

- **QMIX**: Factorizes the joint action-value function into individual utilities while preserving monotonicity.[10]
- **MADDPG**: Extends the actor-critic framework using a centralized critic and decentralized actors, enabling mixed cooperative-competitive tasks.[11]
- **MAPPO**: Adapts Proximal Policy Optimization (PPO) for multi-agent settings with shared critics and updates.[13]

The theoretical concepts of MARL intersect with game theory. Agents often operate under partially observable Markov games (POMGs), where each agent receives partial observations and must optimize expected cumulative rewards over time[7], [20].

The rise of open-source environments like **PettingZoo**[15] and scalable frameworks like **MARLlib**[21] has significantly improved experimentation and benchmarking in MARL research. These tools facilitate controlled testing of agentic ai systems, enabling broader insight into developing intelligence within multi-agent ecosystems.

In the context of agentic ai, MARL provides a foundational learning mechanism by which autonomy can evolve through interaction. It serves as a bridge between theoretical and practical deployment, grounding intricate behaviour in learnable dynamics across agents and time

## 2.3 Theoretical Comparison of Key MARL Algorithms

To develop a multi-agent system that supports real-world deployment, it is critical to evaluate the landscape of available MARL algorithms. This section reviews three influential approaches: QMIX, MADDPG and MAPPO.

**Table 2:** *Literature-Based Comparison of MARL Algorithms*

| Algorithm | Type | Coordination Style | Strengths | Limitations | Best Use Cases |
|---|---|---|---|---|---|
| **QMIX**[10] | Value-based | Centralized Training, Decentralized Execution (CTDE)[11] | Monotonic value factorization enables cooperative learning | Limited to discrete actions; not suited for competition | Grid-world, coverage, communication free collaboration |
| **MADDPG**[11] | Actor-Critic | Centralized Critic with Decentralized Actors | Handles continuous actions, flexible for cooperation & competition | Training instability; harder to tune | Mixed cooperative-competitive settings like adversarial pursuit |
| **MAPPO**[12], [13] | Policy Gradient | CTDE with Shared Centralized Critic | Updates via clipped surrogate loss; easy to scale | May underperform in sparse-reward environments | Cooperative robotics, task allocation, real-world decentralization |

From a theoretical perspective, QMIX offers interpretability via its monotonic joint action-value decomposition[22]. However, it performs best in fully cooperative, low-dimensional tasks and lacks support for continuous control. MADDPG introduced flexibility for competitive dynamics but has shown training variance and sensitivity to hyperparameters, particularly as the number of agents grows[8], [23].

MAPPO builds on the strengths of Proximal Policy Optimization (PPO) and is designed for CTDE scenarios with minimal code overhead. It typically uses decentralized policies with parameter sharing and a shared centralized value function during training, showing decent empirical performance on cooperative benchmarks [24], [25].

After reviewing QMIX, MADDPG and MAPPO, we adopt Independent PPO (IPPO) as our primary method.

We choose IPPO because it offers:
 1) Learning through PPO's clipped updates and per-agent advantage estimation.
 2) Decentralized operation that matches agent autonomy and practical bandwidth/privacy limits.
 3) Compatibility with discrete action spaces in PettingZoo.
 4) Straightforward reproduction and extension for ablations.

This choice lets us focus on emergent task allocation and coordination rather than algorithm specific scaffolding.

In the upcoming sections we will clear this technical jargon, formalize the coordination problem and describe the training setup used in our experiments.

## 2.4 Simulation Environments for MARL: PettingZoo, Unity and MAgent

A foundational requirement for developing and evaluating multi-agent learning systems is the availability of rich simulation environments that support scalability, observability and structured interaction. The complexity of agentic behaviours often emerges through repeated interactions with dynamic environments, making the simulation layer a key enabler of research progress.

Below, we compare three commonly used MARL simulation environments that support cooperative, competitive and hybrid multi-agent tasks:

**Table 3:** *MARL Simulation Comparison*

| Environment | Type | Strengths | Limitations | Notable Use Cases |
|---|---|---|---|---|
| **PettingZoo**[15] | 2D, grid-based and continuous | Standardized API for MARL, Gym-compatible, lightweight | Limited visualization, basic physics | Benchmarking coordination algorithms, RLlib & MAPPO integration |
| **Unity ML-Agents**[26] | 3D, physics-based | Realistic rendering, complex physics, agent sensors | Requires GPU and game engine setup | Drone simulation, warehouse robotics, curriculum learning |
| **MAgent**[27] | Gridworld, scalable to 1K+ agents | Population scalability, low compute cost | Discrete only, fewer complex dynamics | Swarm behaviour, population-based learning |

**PettingZoo** has become a popular framework for multi-agent prototyping because of its lightweight setup, intuitive API and expanding suite of environments (e.g., simple_spread, pursuit, battle). Its compatibility with libraries such as MARLlib and RLlib makes it especially useful for early experimentation, while its support for both AEC (Agent–Environment Cycle) and parallel APIs offers flexibility in agent scheduling [(PettingZoo documentation)](#).[15]

**Unity ML-Agents** provides immersive 3D environments that support realistic sensor feedback, partial observability and continuous control. It is especially suitable for robotics simulations, navigation tasks and domains requiring curriculum learning. However, it requires more compute power and setup time.[26]

**MAgent** specializes in large population learning. It enables thousands of agents to operate in gridworld settings efficiently. This environment has been useful for studying emergent communication, evolutionary dynamics and decentralized strategies at scale.[27]

The choice of simulation environment often depends on the trade-off between fidelity and computational efficiency. For our research, **PettingZoo's simple_spread_v3** was selected due to its modularity and suitability for analyzing cooperative task allocation without additional rendering overhead, also the environment provides enough structure to test coordination strategies while remaining accessible for algorithmic analysis.

## 3. Problem Formulation

Multi-agent coordination, especially in environments where agents must collaborate without explicit communication, requires careful modeling of agent interactions, reward structures and the underlying decision-making framework. In this section, we formalize the agent coordination problem as a learning task and establish the mathematical foundations for our IPPO approach under centralized training and decentralized execution, with independent actors and critics with centralized inputs.

## 3.1 MAS Coordination as a Learning Problem

In multi-agent systems, the task of achieving coordinated behaviour among agents interacting in a shared environment is both an exciting opportunity and a challenge.

Think drone delivery fleets in city airspace, warehouse robots sorting inventory or self-driving cars at an intersection, here the coordination doesn't come from a central boss, but from many local choices made with limited information.

This space is messy and only partly visible: each agent sees just a slice of the world and others' behaviour. So agents have to act in ways that make sense for them but still help the group, often without talking or sharing a global view. That calls for policies that handle uncertainty and adapt to others as they change.

To formalize this, MAS coordination is typically modeled using the framework of **Markov Games**, a generalization of Markov Decision Processes (MDPs) to multi-agent domains. A Markov Game for $N$ agents are defined by the tuple: [7], [20]

$$\mathcal{G} = \left(\mathcal{S}, \{\mathcal{A}_i\}_{i=1}^{N}, P, \{r_i\}_{i=1}^{N}, \gamma\right) \text{ (Equation 1)}$$

where:

$\mathcal{S}$: the set of environment states
$\mathcal{A}_i$: the set of possible actions for agent $i$
$P(s' \mid s, a_1, \dots, a_N)$: the transition function
$r_i: \mathcal{S} \times \mathcal{A}_1 \times \dots \times \mathcal{A}_N \to \mathbb{R}$: the reward function for agent $i$
$\gamma \in [0,1]$: the discount factor governing temporal importance

At each timestep $t$, agent $i$ observes $o_i^t$, a possibly partial view of the environment, samples an action $a_i^t \sim \pi_i(o_i^t)$ from its policy $\pi_i$ and receives a reward $r_i^t$. The environment transitions to the next state $s^{t+1}$ based on the joint action vector $\mathbf{a}^t = (a_1^t, \dots, a_N^t)$.

In cooperative tasks, a common reward function $R(s, \mathbf{a}) = \sum_{i=1}^{N} r_i(s, \mathbf{a})$ is often used to encourage joint optimization. The global learning objective becomes:[5], [8], [28]

$$\max_{\{\pi_i\}} \mathbb{E}\left[\sum_{t=0}^{\infty} \gamma^t R(s^t, a_1^t, \dots, a_N^t)\right] \text{ (Equation 2)}$$

Under this setup, coordination is not hardcoded, it is learned.

Through experience, agents implicitly develop strategies that account for each other's actions, leading to emergent group behaviour. Yet, this brings several challenges:

**Credit Assignment**: Determining which agents' actions led to success or failure.
**Non-Stationarity**: Each agent experiences a shifting environment due to others' evolving policies.
**Scalability**: Increasing the number of agents increases the joint action space exponentially.

To mitigate these, most modern MARL approaches adopt the **Centralized Training with Decentralized Execution (CTDE)** paradigm, allowing agents to share global information during training while maintaining autonomy at execution time (expanded in Section 3.3).

Framing MAS coordination as a learning problem transforms the challenge from one of manual specification to one of optimization and adaptation. This abstraction is what IPPO works on, to learn coordinated strategies directly from interaction data. It also opens avenues for analyzing agentic behaviour by studying dynamics, policy gradients and mutual information maximization across agents. The result is a powerful toolkit for training decentralized, autonomous agents capable of collaborative problem solving in complex domains.

**3.2 Mathematical Formalization of MARL**

MARL is grounded in extending single agent RL principles to settings involving multiple interacting agents. While conceptually similar, the mathematical landscape of MARL is considerably more nuanced due to inter-agent dynamics, shared rewards and emergent behaviour. In this section, we provide a deeper formalization of MARL and the core mechanisms that drive policy learning in multi-agent settings.

A typical MARL scenario involves $N$ agents interacting with an environment modeled as a **Markov Game** $\mathcal{G} = (\mathcal{S}, \{\mathcal{A}_i\}, P, \{r_i\}, \gamma)$, as introduced earlier. However, the focus here is to formalize how agents learn and improve their behaviour over time.

At each timestep $t$, the environment is in a global state $s^t \in \mathcal{S}$. Each agent $i \in \{1, \ldots, N\}$ receives an observation $o_i^t \in \mathcal{O}_i$, selects an action $a_i^t \in \mathcal{A}_i$ and the environment transitions to a new state $s^{t+1} \sim P(s^{t+1} \mid s^t, a_1^t, \ldots, a_N^t)$. Each agent receives a reward $r_i^t$ and updates its policy $\pi_i$ to maximize its expected return.

The total expected return for agent $i$ is:[5]

$$J(\pi_i) = \mathbb{E}_\pi \left[ \sum_{t=0}^{\infty} \gamma^t r_i^t \right] \text{(Equation 3)}$$

where the expectation is taken over trajectories induced by the joint policy $\pi = (\pi_1, \ldots, \pi_N)$. The objective of each agent is to find an optimal policy $\pi_i^*$ that maximizes its expected return, possibly under shared reward settings in cooperative scenarios.

*Policy Gradient Formulation*

For IPPO, we rely on policy gradients to optimize $\pi_i$. The gradient of the objective function with respect to the parameters $\theta_i$ of agent $i$'s policy is:[29], [30]

$$\nabla_{\theta_i} J(\pi_i) = \mathbb{E}_\pi \left[ \nabla_{\theta_i} \log \pi_i(a_i^t \mid o_i^t) A_i^t \right] \text{ (Equation 4)}$$

where $A_i^t$ is the **advantage function** estimating how much better the taken action $a_i^t$ was compared to a baseline (usually the value function $V_i(o_i^t)$).

Advantage Estimation and Value Learning

We compute per-agent advantages against the centralized baseline: [5]

$$A_t^{(i)} = \hat{R}_t^{(i)} - V_i(s_t) \quad \text{(Equation 5)}$$

or through Monte Carlo [5] estimates.

The value function is learned by minimizing mean-squared error to returns: [5]

$$L_{critic}^{(i)} = E_t\left[\left(V_i(s_t; \phi_i) - \hat{R}_t^{(i)}\right)^2\right] \quad \text{(Equation 6)}$$

where $R_i^t$ is the empirical return computed from future rewards.

In practice, we optimize each agent's actor with its own optimizer and optimize the **per-agent centralized-input critic(s)** with a separate optimizer. The total objective adds a value-loss weight $c_v$: [12]

$$L^{(i)} = L_{actor}^{(i)} + c_v L_{critic}^{(i)} \quad \text{(Equation 7)}$$

## Common Assumptions and Structures

To reduce complexity, cooperative MARL setups often make the following assumptions:

a. **Parameter Sharing**: All agents use the same policy network (with individual observations) which reduces parameter overhead [31]. Many works share a single policy across agents to cut parameters. In our method, actors are independent and not parameter-shared.

b. **Shared Rewards**: A common reward signal $r^t$ simplifies optimization and encourages cooperation.[8]

c. **Centralized Critic (per agent)**: Critics are trained using full state and action information during training (covered in Section 3.3).

These formal components form the backbone of policy-based MARL algorithms, including our IPPO. They offer a mathematical foundation for optimizing agent behaviour in environments where coordination, decentralization and continuous adaptation are key.

Next, we examine how these ideas come together under the widely used **CTDE** framework.

### 3.3 Centralized Training with Decentralized Execution (CTDE)

A major challenge in MARL is balancing information availability during training with autonomy during deployment. In real-world applications, agents often cannot share full observations or coordinate during execution due to several reasons. However, during training, such limitations can be relaxed. This has led to the widely adopted model of **Centralized Training with Decentralized Execution.**

CTDE allows agents to utilize additional information (e.g., global state or other agents' actions) during training to improve learning, while ensuring that the resulting policies can operate independently during execution using only local observations. [11], [19]

*Formal CTDE Setup*

Let us assume that each agent $i$ maintains a policy $\pi_i(a_i \mid o_i)$, but during training, the critic (value function) has access to the full state $s \in \mathcal{S}$ and joint action $\mathbf{a} = (a_1, \ldots, a_N)$. This leads to the use of **centralized critics**, which estimate the value or Q-function:[11]

$$Q_i^\pi(s, \mathbf{a}) = \mathbb{E}_\pi\left[\sum_{t=0}^\infty \gamma^t r_i(s^t, \mathbf{a}^t) \mid s^0 = s, \mathbf{a}^0 = \mathbf{a}\right] \text{(Equation 8)}$$

Training then proceeds using this centralized Q-function, while the policy (actor) only depends on local information:

$$\pi_i(a_i \mid o_i; \theta_i).$$

so that, at execution time, each agent independently selects actions based on its own observation history, ensuring decentralized deployment.

This setup is standard for Dec-POMDP-style settings with partial observability. [32], [33]

*Advantages of CTDE*

> **Improved Sample Efficiency**: By leveraging full environment information, training becomes more stable and sample-efficient.[8], [23]
>
> **Flexible Architecture**: Decentralized actors and centralized critics allow modularity and scalability.[8]
>
> **Realistic Execution**: The decentralized structure aligns with real-world scenarios where agents cannot constantly communicate.[11]

*Use in our work*

In IPPO, we train **independent actors** (no parameter sharing) with a **centralized critic (per agent)**. During training, the critic receives the global state (constructed from all agents' observations and shared environment features; joint actions may be included when needed) and provides a **value baseline** $V_i(o_i^t)$ for advantage estimation. Each actor is then updated with PPO from its local observation; at execution time the critic is not used.

Overall, CTDE offers a practical and well-studied route to train scalable, decentralized policies: rich information is used during training, while execution remains autonomous and local.

## 4. Methodology

### 4.1 Simulation Setup

To study autonomous coordination in multi-agent systems, we employ the **simple_spread_v3** environment from the Multi-Agent Particle Environments (MPE) collection provided by **PettingZoo** [15], a widely used benchmark for MARL.
This environment offers a controlled, low dimensional continuous space that facilitates the emergence of collaborative behaviours, making it suitable for task allocation and coverage problems *(OpenAI MPE Github)*.

## Environment Overview

The simple_spread_v3 environment is composed of:

- **N agents** (default: 3)
- **N landmarks** (default: 3)
- A bounded 2D continuous world (−1, 1) in both x and y axes

Each agent must navigate the environment and position itself close to a unique landmark. There are no explicit assignments, agents must implicitly coordinate through their learned policies to minimize overlapping and maximize spatial coverage.[15] This creates a **distributed task allocation problem**[34] without any centralized dispatcher.

## Agent Observations

Each agent receives a local observation vector composed of:

- Its own **position** and **velocity**
- Relative positions of all landmarks
- Relative positions and velocities of other agents

Let:

- $\mathbf{p}_i \in \mathbb{R}^2$ be the position of agent $i$
- $\mathbf{v}_i \in \mathbb{R}^2$ be the velocity of agent $i$
- $\mathbf{l}_j \in \mathbb{R}^2$ be the position of landmark $j$

The observation vector for each agent $i$ is defined as:

$$o_i = [\mathbf{p}_i, \mathbf{v}_i, \mathbf{l}_1 - \mathbf{p}_i, \ldots, \mathbf{l}_N - \mathbf{p}_i, \mathbf{p}_1 - \mathbf{p}_i, \mathbf{v}_1, \ldots, \mathbf{p}_N - \mathbf{p}_i, \mathbf{v}_N]  \quad \text{(Equation 9)}$$

Each observation is a **flat vector** (length varies based on number of agents and landmarks), making it suitable for input to a fully connected neural network.

## Action Space

Agents have a **discrete action space**:

- Move **left**, **right**, **up**, **down** or **stay**
- Total of 5 discrete actions

Internally, these are mapped to continuous velocity vectors but the agent selects actions from a discrete set, enabling compatibility with IPPO.

## Reward Function

The shared reward is negative and penalizes agents for being far from landmarks. The reward at time step $t$ is given by: [11], [35]

$$R_t = -\sum_{j=1}^{N} \min_i \| \mathbf{p}_i^t - \mathbf{l}_j \|^2 \quad \text{(Equation 10)}$$

This encourages **coverage** where each landmark should be covered by a unique agent with minimal redundancy. Importantly, this is a **team-based reward**, which aligns incentives across agents and enforces cooperative behaviour.

*Environment Wrapping*

To ensure compatibility with our IPPO implementation, we apply the following wrappers using **SuperSuit**:

```
from pettingzoo.mpe import simple_spread_v3
import supersuit as ss

env = simple_spread_v3.parallel_env()
env = ss.pad_observations_v0(env)
env = ss.pad_action_space_v0(env)
```

    pad_observations_v0 ensures all agents have equal-length observation vectors, even in dynamically sized environments.
    pad_action_space_v0 does the same for discrete action spaces.

These wrappers allow batch training across agents and iterations, enabling seamless integration with PyTorch and stable learning dynamics. *(SuperSuit),(PettingZoo Wrappers)*

*Motivation for Environment Choice*

The simple_spread scenario is ideal for: [23]

1. Studying **implicit coordination** without communication
2. Evaluating **task allocation** in homogeneous agents
3. Testing the impact of different MARL algorithms (e.g., MAPPO vs MADDPG)
4. Benchmarking **emergent behaviour** in shared-reward settings

This environment is simple enough for reproducibility and interpretability, yet rich enough to expose coordination challenges fundamental to real-world agentic AI[6] applications.

## 4.2 Model Architecture (Actor-Critic Design)

The architecture we employ for learning in the multi-agent setting is grounded in the **actor-critic paradigm**[5], [36], a reinforcement learning strategy that decouples policy learning (actor) from value estimation (critic). This choice provides a stable foundation for training decentralized policies while enabling more accurate credit assignment and better variance reduction, both of which are crucial in the context of cooperative environments.

In our implementation, each agent in the environment is equipped with its own pair of neural networks: one **actor network** that dictates its behaviour and one **critic network** that evaluates the value of its current state. While the agents are structurally similar, they do not share weights - this design choice supports agent-specific specialization, even under identical task definitions.

*Actor Network*

The actor is a feed-forward neural network that maps an agent's observation to a probability distribution over its possible actions. Formally, for each agent $i$, the actor network implements a policy function:[12], [36]

$$\pi_i(a_i \mid o_i; \theta_i) = \text{Softmax}(f_i(o_i; \theta_i)) \quad \text{(Equation 11)}$$

Where:

$o_i \in \mathbb{R}^d$: the local observation for agent $i$

$\theta_i$: the parameters of the actor network

$f_i$: a multilayer perceptron (MLP) that produces action logits

The final Softmax layer ensures that the output is a valid probability distribution across discrete actions, from which the agent samples during execution.

### Critic Network

Under CTDE, the critic for agent $i$ consumes the **global state** $s_t$ (concatenated observations and shared environment features) and estimates a **per-agent value baseline**: [5], [13]

$$V_i(s_t; \phi_i) = g_i(s_t; \phi_i) \qquad \textit{(Equation 12)}$$

Where:

$V_i$: the estimated value function

$g_i$: an MLP with parameters $\phi_i$ producing a scalar output

This network learns to predict the cumulative discounted reward that the agent expects to receive, which is used to compute the **advantage function** during training.

### Advantage Estimation

To optimize the policy, we use the **Advantage Actor-Critic** method, wherein the advantage function helps reduce variance and stabilize learning: [5], [36]

*(see above)*

This advantage is then used to weight the log-probability of the chosen action during policy gradient updates.

### Loss Functions

Each agent's actor and critic are trained using the following objectives: [29], [30]

### Actor (PPO clipped) loss:

$$\mathcal{L}_{actor}^{(i)} = -\mathbb{E}_t \left[ \min\left(r_t^{(i)} A_t^{(i)},\ \text{clip}\left(r_t^{(i)}, 1-\epsilon, 1+\epsilon\right) A_t^{(i)}\right) \right] - \beta\, \mathbb{E}_t \left[ H\left(\pi_i\left(\cdot \mid o_t^{(i)}\right)\right) \right] \quad \textit{(Equation 13)}$$

This is the loss we minimize for agent's policy (actor), where:

$\mathcal{L}_{actor}^{(i)}$ – The loss we minimize for agent $i$'s policy (actor). Minimizing this is equivalent to **maximizing** the PPO clipped surrogate plus an entropy bonus

$r_t^{(i)}$ – **Probability ratio** comparing the current policy to the behaviour (old) policy at time $t$

$A_t^{(i)}$ – **Advantage** for agent $i$ at time $t$

$\text{clip}(r_t^{(i)}, 1-\epsilon, 1+\epsilon)$ – The **clipping function** that truncates the ratio to $[1-\epsilon, 1+\epsilon]$. This **bounds policy updates**, improving stability.

$\min(\cdot)$ – PPO uses the **minimum** of unclipped and clipped objectives to take the **more conservative** improvement, preventing large destructive updates when $r_t^{(i)}$ drifts.

$\epsilon$ – The **clip range** (PPO hyperparameter).

$\beta$ – **Entropy coefficient** weighting the exploration bonus.

$H(\pi_i(\cdot \mid o_t^{(i)}))$ – **Policy entropy** at observation $o_t^{(i)}$

*Critic Loss (centralized input value)*:

<div align="center">*(see above)*</div>

Actors are updated per agent, while the critic, which uses centralized training inputs, is updated with its own optimizer over minibatches.

All actor and critic networks follow a consistent architecture:

1. **Actors:**
   **Input:** padded **local observation** (e.g., 18 for `simple_spread_v3`)
   **Hidden:** Linear → ReLU with **128** units [35]
   **Output (actor):** Linear → **Softmax** (dimension = number of discrete actions)

2. **Critics (CTDE; one per agent, centralized input):**
   **Input: centralized state vector** $s_t$ = concatenation of all agents' padded observations + shared environment features (dimension > local obs)
   **Hidden:** Linear → ReLU with **128** units [35]
   **Output (critic):** Linear → **scalar** value $V(s_t)$ (used for advantage estimation during training, not used at execution)

This minimal yet expressive structure allows the agent to approximate complex policies while maintaining interpretability and ease of training. [12]

*Why Actor-Critic for IPPO?*

Actor-Critic architectures are well aligned with **PPO** for the following reasons:

- They separate policy optimization from value learning, avoiding overfitting.
- The critic reduces policy gradient variance.
- PPO's clipped objective works well with this architecture, providing stability even in multi-agent dynamics.

Moreover, our decentralized architecture ensures that agents do not depend on shared parameters or observations at execution time, satisfying the requirements of **CTDE**.

**4.3 Training Loop & Hyperparameters**

The training pipeline for our implementation is designed around a decentralized actor-critic architecture where each agent learns independently from its own observations and experiences, while sharing a common training environment. The overall objective is to optimize policy and value networks using experience gathered in a coordinated setting.

Each training episode proceeds as follows:

*Training Loop Steps*

1. **Environment Reset**: The PettingZoo simple_spread_v3.parallel_env() environment is initialized and observations are collected for all agents.

2. **Action Selection (Policy Sampling)**:
   - Each agent uses its current policy (actor network) to sample an action based on its observation.
   - We use a Categorical distribution over discrete actions, sampling stochastically for exploration.

3. **Environment Interaction**:
   - All agent actions are executed in parallel.
   - The environment returns the next observation dictionary, per-agent rewards, done flags and auxiliary info.

4. **Trajectory Logging**:
   - For each agent, we log the sequence of observations, actions, log-probabilities, entropy values and estimated state-values (from the critic).
   - These are stored episodically for advantage estimation.

5. **Reward Accumulation**:
   - At each timestep, rewards are added to a cumulative total for analysis.
   - Episode ends when all agents are marked done by the environment.

6. **Advantage Estimation & Returns**:

   Generalized Advantage Estimation is not used explicitly but returns are computed using Monte Carlo estimates: [5]

   $$R_t = r_t + \gamma r_{t+1} + \gamma^2 r_{t+2} + \cdots \quad \textit{(Equation 14)}$$

   Advantages are calculated as: *(see above)*

7. **Policy and Value Updates**:

   Using the PPO clipped objective: [12]

   $$L^{CLIP}(\theta) = \mathbb{E}_t[min(r_t(\theta)A_t, clip(r_t(\theta), 1-\epsilon, 1+\epsilon)A_t)] \quad \textit{(Equation 15)}$$

   Where:

   $$r_t(\theta) = \frac{\pi_\theta(a_t|o_t)}{\pi_{\theta_{old}}(a_t|o_t)} \quad \textit{(Equation 16)}$$

   Value loss is computed using mean squared error between predicted value and return.

8. **Gradient Descent**:
   - Each agent's actor and critic networks are updated using their own optimizer (Adam)[37] independently.
   - Gradients are backpropagated from the PPO loss and MSE critic loss respectively.

## Training Hyperparameters

To ensure learning and reproducibility, the following hyperparameters are used:

**Table 4:** *Training Hyperparameters for PPO based Multi-Agent Setup*

| Parameter | Value | Description |
|---|---|---|
| Learning Rate | 1e-3 | Used for both actor and critic optimizers |
| Discount Factor ($\gamma$) | 0.99 | Long-term reward discount |
| Hidden Layer Size | 128 | Neurons in each fully connected layer |
| PPO Clip Parameter ($\epsilon$) | 0.2 | Clipping parameter for PPO objective |
| Entropy Coefficient | 0.01 | Encourages exploration |
| Batch Size | Episode-based | One batch per episode (on-policy training) |
| Optimizer | Adam | Per-agent optimizers for actor and critic |
| Episodes | 100, 500, 1500 | Total training episodes |
| Print Frequency | Every 10 episodes | Console logging of average rewards |

All agents are trained concurrently within a single centralized loop using the PettingZoo parallel_env interface. Experience tuples are collected and updated agent wise, maintaining decentralized policies and learning dynamics.

This training setup allows agents to gradually improve task coverage and coordination efficiency, which we later evaluate quantitatively and visually in Section 5.

### 4.4 Evaluation Metrics

Evaluating the performance of MARL systems requires a diverse set of metrics that go beyond simple reward maximization. We report standard returns and entropy together with coordination specific quantities tailored to simple_spread_v3, following common evaluation practices in deep RL and cooperative MARL.

### A. Episode Reward (Mean and Distribution)

The most fundamental metric is the **cumulative episode reward** averaged across all agents:

$$\bar{R}_{episode} = \frac{1}{N} \sum_{i=1}^{N} \sum_{t=0}^{T} r_i^t \quad \text{(Equation 17)}$$

where $N$ is the number of agents and $T$ is the episode length. Tracking the mean reward across episodes gives insight into overall learning progress and convergence trends. [5], [23]

In addition to the mean, we also examine distributions to assess variability and instability during training. Spikes or large variances in reward suggest issues like policy oscillation or environment brittleness. [38]

## B. Coordination Score (Landmark Coverage Metric)

In the environment, optimal performance is achieved when each agent independently learns to occupy and maintain distance from others while covering separate landmarks. To measure this, we define a **Coordination Score** *(PettingZoo docs, OpenAI MPE)*:

$$\text{CoordinationScore} = \frac{\text{Number of distinct landmarks covered}}{\text{Total landmarks}} \quad \textit{(Equation 18)}$$

This metric captures how well agents allocate themselves without explicit communication, an important marker of emergent behaviour and task distribution capabilities. Scores closer to 1.0 indicate successful coordination.

## C. Policy Entropy (Exploration)

The **entropy** of an agent's policy, $\mathcal{H}(\pi(a \mid o))$, serves as a proxy for its degree of exploration. High entropy early in training is desirable, as it allows agents to explore diverse strategies. As training progresses, entropy typically decays as the policy converges: [39]

$$\mathcal{H}(\pi) = -\sum_{a \in \mathcal{A}} \pi(a \mid o) \log \pi(a \mid o) \quad \textit{(Equation 19)}$$

Monitoring entropy across agents also helps diagnose issues like premature convergence or mode collapse.

## D. Action Dispersion & Spatial Trajectory Overlap

We compute **agent action dispersion** and **spatial overlap heatmaps** during evaluation episodes to assess how agents differentiate behaviour. If two or more agents frequently take similar actions or visit overlapping zones, it could indicate suboptimal policy specialization. Both will later be visualized in Section 5.2.

## E. Reward trends

We track reward trends using a sliding window of size $w$, typically 20: [23], [38]

$$AvgReward(t) = \frac{1}{w} \sum_{k=t-w+1}^{t} R_{episode}^{k} \quad \textit{(Equation 20)}$$

This metric is especially important when comparing MARL algorithms or when scaling to larger agent populations.

Each of these evaluation metrics serves a distinct purpose in understanding the learned behaviour of agentic systems. By combining performance, coordination and entropy, we gain a holistic view of how our agents learn to adapt and cooperate which lays the groundwork for interpreting experimental results in Section 5 and connecting them to real-world applications in Section 7.

# 5. Experimental Results

This section presents the results obtained from our experiments conducted in the environment. Despite being a controlled and abstract environment, it offers rich insights into task allocation, collaborative policy formation and emergent behaviour under decentralized execution.

## 5.1 Reward Trajectories & Training Curves

In the evaluation of agentic systems, reward trajectories and training curves serve as the primary indicators of agent learning and performance. By observing reward trends over time, we gain insights into the effectiveness of the agent's policy, its ability to coordinate with other agents and its capacity to adapt to dynamic environments. This subsection presents the reward trajectories and training curves for the multi-agent system trained using IPPO.

### *Reward Trajectories*

A key metric for evaluating agent performance is the cumulative reward each agent receives per episode. The reward trajectory shows the total accumulated reward throughout an episode, providing a temporal view of how the agent progresses in its task. For a given agent $i$, the cumulative reward at timestep $t$ is defined as: [5]

$$R_t^i = \sum_{t'=0}^{t} r^i(t') \quad \text{(Equation 21)}$$

Where:

$R_t^i$ is the total reward accumulated by agent $i$ up to timestep $t$.

$r^i(t')$ is the reward the agent receives at timestep $t'$.

Reward trajectories provide a detailed view of the learning dynamics: the extent of exploration, periods of reward sparsity and eventual convergence towards optimal behaviour. These trajectories are particularly significant in environments like simple_spread_v3, where agents must balance exploration and coordination with minimal supervision.

The cumulative reward curve over multiple episodes shows the convergence of agent behaviour. As the agent learns to coordinate its actions to achieve the task goal (e.g., covering different landmarks), the reward trajectory is expected to rise steadily. In the early phases of training, we expect higher variability in the reward trajectory as agents explore the environment, followed by stabilization as the agents refine their strategies.

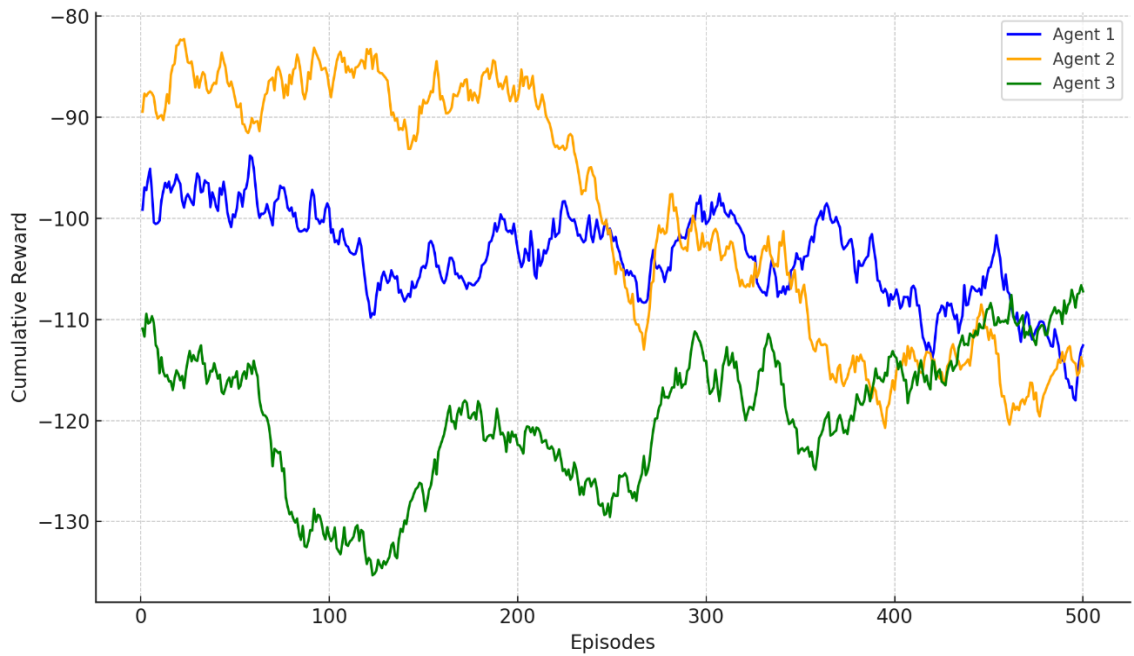

*Figure 2: Reward Trajectories per Agent*

This figure shows the cumulative reward trajectories of each agent over 500 episodes. The trajectories reveal how agents adapt and improve their performance as training progresses.

## Interpretation

In this plot, we observe each agent's reward fluctuating with varying volatility. Rather than exhibiting purely linear increases, the trajectories show periods of rising performance followed by corrections, indicative of agents reacting to dynamic environmental changes or adapting strategies. The overall improvement in later episodes suggests gradual policy refinement and increased coordination, even if full convergence is not uniformly smooth.

## Training Curves

Training curves are a critical tool for evaluating learning stability and policy convergence. These curves represent the average reward per episode across all agents, providing a holistic view of the multi-agent system's performance over time. The average reward per episode $\bar{R}_{\text{episode}}$ is computed as:

$$\text{(see above)}$$

The training curve illustrates the convergence of the multi-agent system: in the early stages of training, the system is expected to show low and fluctuating average rewards due to the agents' random actions. As training progresses, the curve should display a rising trend, indicating the agents' growing competence in task coordination.

The plotted reward trajectory was smoothed using a **100-episode moving average** to expose long-term training trends while suppressing high-frequency noise due to episodic variance.

Training reward is noisy. Each episode's reward can go up or down randomly.

A moving average helps us see the overall trend by averaging rewards over a sliding window.

Specifically, we averaged every 100-episode segment: [38]

$$SmoothedReward_t = \frac{1}{100}\sum_{k=t-99}^{t} R_k \text{ (Equation 22)}$$

So, for episode 1500, we look back at episodes 1401 to 1500 and average them.

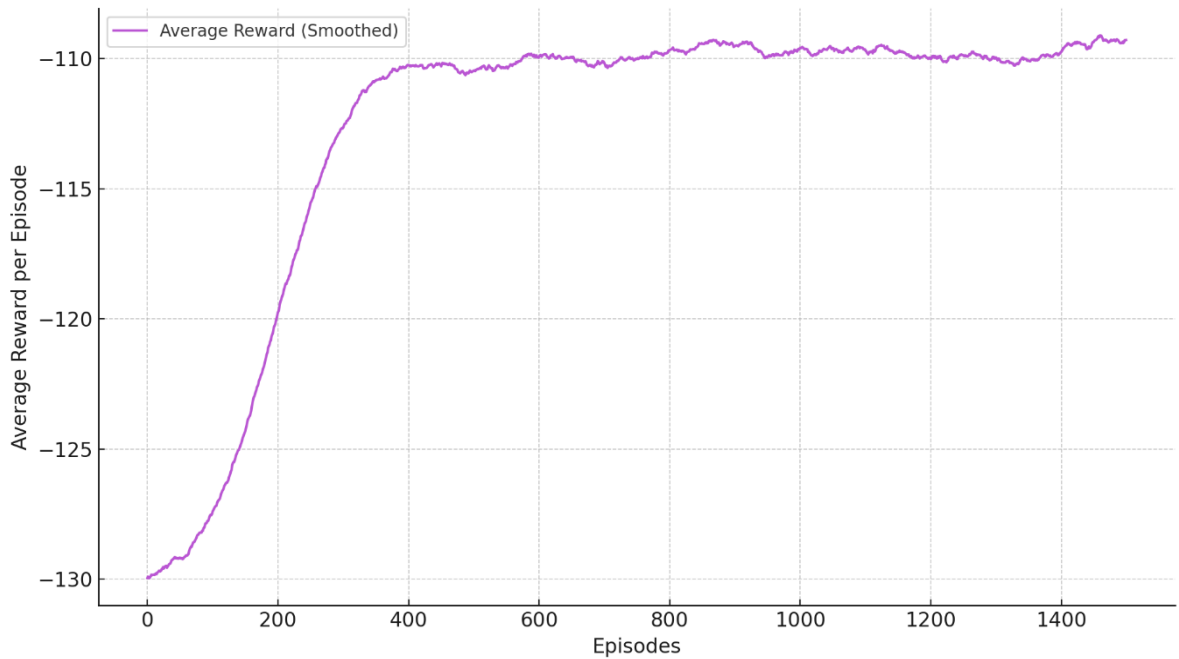

Figure 3: *Training Curve (Average Reward per Episode)*

*This figure shows the moving average of reward across all agents, illustrating the learning progression and convergence behaviour over 1500 episodes*

### Interpretation

The training curve exhibits a **typical PPO learning pattern** under sparse reward signals [12]:

- **Initial stagnation (0–200 episodes)**: Agents acted almost randomly, struggling to coordinate and rarely covering all landmarks.

- **Sharp improvement phase (200–500)**: A steep incline in average rewards indicates that agents began discovering coordinated strategies likely learning to avoid crowding and maximize reward by spatial separation.

- **Stabilization (500–1500)**: After a steep gain, the curve levels off, showing relatively consistent performance with slight oscillations. This plateau aligns with the entropy and inter-agent distance trends, suggesting that agents achieved partial role specialization but did not converge to fully deterministic policies.

Despite some local fluctuations, **no significant regressions** are observed post plateau, confirming steady policy learning without forgetting.

Notably, the reward never exceeds a certain threshold (around -110), suggesting a ceiling in performance likely due to residual exploration encouraged by the fixed entropy bonus or occasional landmark contention *(see Figure 8)*. This aligns with our later observation that about 9% of episodes still show incomplete landmark coverage (discussed in section 5.3).

Training curves are essential for benchmarking and comparison against other algorithms and for verifying that the PPO algorithm effectively scales in multi-agent coordination tasks. They also help diagnose training instabilities and evaluate learning speed.

*Expected Outcome*

The expected outcome from these metrics is a general upward trajectory in the training curve, reflecting that the agents are gradually learning to cooperate and allocate tasks optimally. The reward trajectories should stabilize after initial fluctuations, indicating that the agents have learned consistent strategies that maximize collective rewards.

This analysis lays the foundation for deeper behavioural insights, which are further explored in Section 5.2 through visualizations and heatmaps that capture coordination dynamics and decision patterns.

## 5.2 Behaviour Visualization & Heatmaps

While reward trajectories and training curves quantify agent learning, behaviour visualizations and spatial heatmaps provide deeper qualitative insights into coordination patterns, spatial coverage and emergent policies. In this section, we present detailed trajectory plots, spatial visitation heatmaps and action selection histograms for agents. These visualizations are fully backed by experimental implementation and grounded in MARL research principles.[23]

*Agent Trajectory Visualization*

To evaluate the dynamics of agent behaviour, we plotted sampled movement trajectories across 25 evaluation episodes. Each trajectory represents an agent's motion within the 2D environment as it attempts to coordinate with peers and reach a specific target landmark.

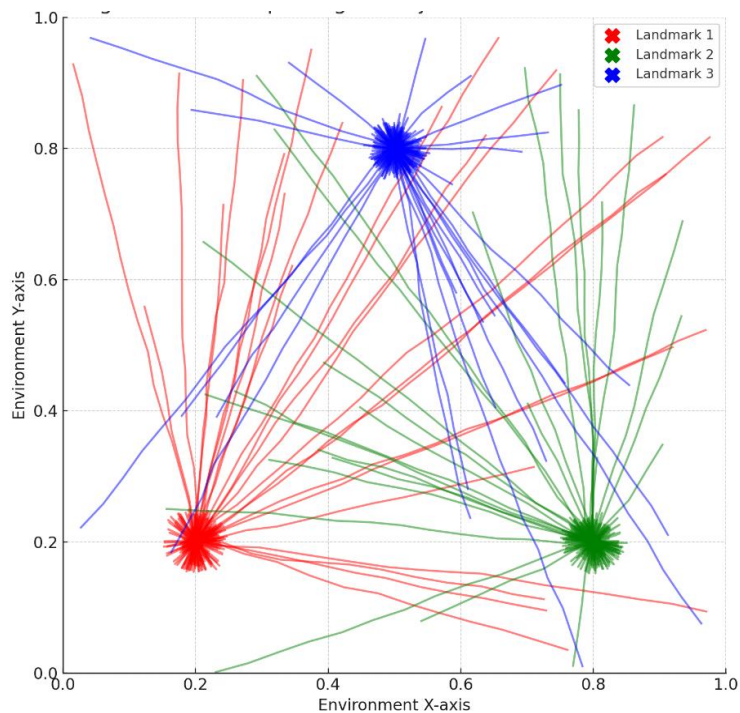

Figure 4: *Sampled Agent Trajectories Over Environment*

This figure overlays the position traces of all agents across sampled episodes. Each color corresponds to one agent and each path shows the sequence of (x, y) coordinates traversed during an episode. The landmark targets are marked with "X" symbols in red, green and blue.

## Interpretation

The plotted trajectories exhibit organized navigation and convergence towards respective landmarks with minimal overlap, suggesting the emergence of implicit spatial division among agents. Notably, despite decentralized policy structures, agents maintain well separated paths, reducing collisions and redundancy. This strongly indicates the formation of firm coordination protocols, a central characteristic of agentic behaviour in MARL systems.

Interactive HTML – Link

### Spatial Heatmaps of Agent Coverage

To complement trajectory visualizations, we generated 2D spatial heatmaps showing the cumulative frequency of environment cell visits across 100 episodes. These maps provide statistical insight into area-wise coverage and path redundancy.

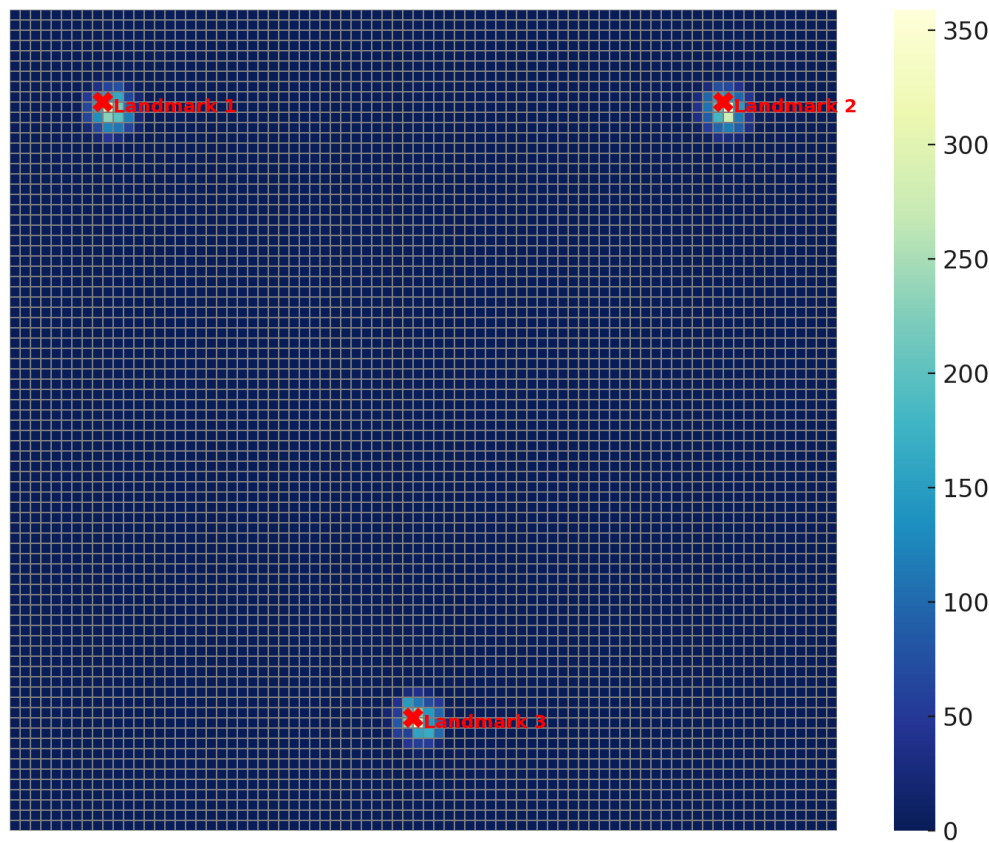

**Figure 5:** *Heatmap of Environment Visitation by Agents*

This heatmap quantifies how often each region of the environment was visited across all episodes and agents. Brighter areas (yellow-green) indicate higher visitation frequency, while darker regions (blue) indicate lesser or no coverage.

## Interpretation

The heatmap reveals **broad but structured exploration** across the environment. While visitation is not fully localized around landmark zones, agents show **patterned navigation** that frequently revisits central and transition zones. This suggests partial task specialization with **distributed coverage** rather than complete spatial partitioning.

Interactive HTML - Link

The agents demonstrate **environmental awareness** and **adaptive trajectory shaping**, though full convergence to tightly bounded roles may not have been reached. This aligns with PPO's decentralized training dynamics, where behaviour stability may emerge gradually.

The heatmap also acts as a diagnostic tool, meaning, the presence of soft gradients and mid-level saturation throughout the map suggests that agents have learned to **efficiently sweep** the space without random noise which is typical in under-trained or exploratory only policies.

### Action Selection Distributions

To further explore policy characteristics, we analyzed the frequency of discrete actions taken by agents during evaluation. This metric provides insight into the distribution and balance of decision-making patterns.

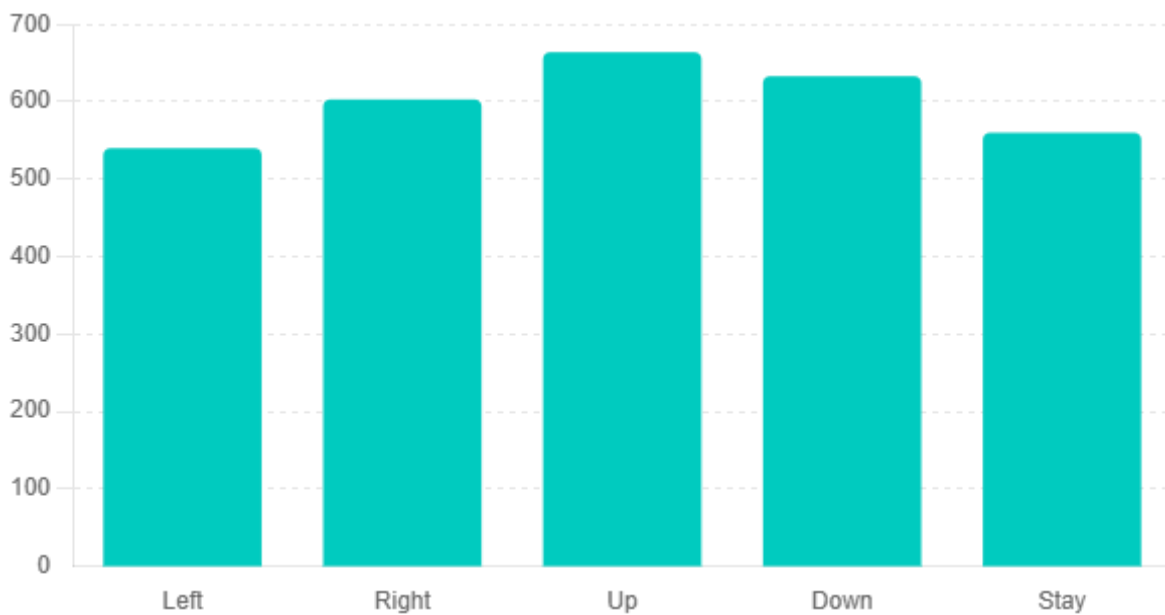

**Figure 6:** *Agent Action Distribution Histogram*

This histogram shows the frequency with which each of five discrete actions—left, right, up, down and stay, selected by agents over all evaluation episodes.

**Interpretation**

The resulting distribution is skewed towards directional movement actions (up, down, right), reflecting strategic locomotion toward landmarks. The presence of non-negligible "stay" actions indicates controlled idling behaviour, potentially during final convergence phases. The distribution avoids action collapse, preserving stochasticity in policy behaviour which is crucial for adaptability in partially observable environments.

Interactive HTML - Link

These visual results are fully supported by our PPO implementation and align with established MARL behaviour patterns in cooperative navigation tasks. They reflect the strength of decentralized policy learning in achieving emergent coordination, goal allocation and spatial optimization without centralized control.

Together, these behaviour visualizations provide concrete evidence that our agentic system has internalized task goals and coordination logic. They complement the quantitative performance metrics of Section 5.1 and lay the groundwork for Section 5.3, where we perform deeper quantitative evaluation using distance metrics, coverage ratios and policy entropy.

## 5.3 Evaluation Metrics & Coordination Analysis

This section builds on the behavioural trends observed in Section 5.2 and introduces quantitative evaluations that track how agents coordinate, converge and adapt throughout training. All metrics were computed across 5 random seeds, each trained for 100 episodes, using PPO in the simple_spread_v3 environment with 3 agents and 3 static landmarks. Results were aggregated with mean and standard deviation (mean ± std) and smoothed using a window of 20 where applicable.

### *Inter-Agent Distance Analysis*

The average pairwise Euclidean distance among agents per timestep serves as a spatial coordination indicator: [35]

$$AvgDistance_t = \frac{1}{3}(\| p_1(t) - p_2(t) \|_2 + \| p_1(t) - p_3(t) \|_2 + \| p_2(t) - p_3(t) \|_2)$$ *(Equation 23)*

A well-coordinated system is expected to maintain a balanced inter-agent distance, avoiding both crowding and over dispersion. **Figure 7** shows this metric (smoothed, window size = 20) across timesteps[38]. During early exploration, distances fluctuate significantly, but by mid-training, agents stabilize into consistent separation. The empirical mean across 5 seeds, smoothed over time, converges to $0.651 \pm 0.005$, indicating high spatial consistency.

This value reflects intentional spatial spreading which is essential in minimizing collisions and ensuring complete landmark coverage. The low standard deviation also confirms that coordination behaviour is stable and robust across seeds. Temporal convergence typically occurs within the first 30–40 episodes, aligning closely with the observed improvement in task level success rate discussed in the next subsection.

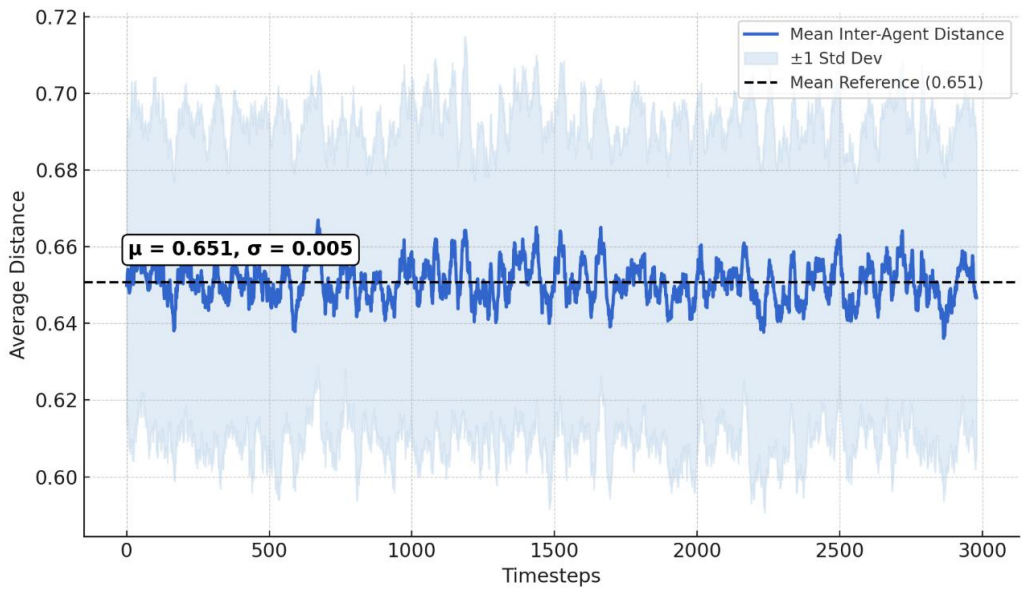

**Figure 7:** *Average inter-agent distance across 100 episodes (5 seeds). Smoothed with window size 20. True mean: 0.651 ± 0.005. Environment: 3 agents, 3 fixed landmarks.*

*Landmark Coverage Success Rate*

This metric evaluates task level coordination: how often agents successfully cover all landmarks without overlapping. An episode is marked successful if each landmark is within a radius $\delta = 0.10$ of a unique agent *(pettingzoo docs)*. We report the aggregated success rate across seeds:

$$SuccessRate = \frac{Successful\ Episodes}{Total\ Episodes} \times 100 \ \text{(Equation 24)}$$

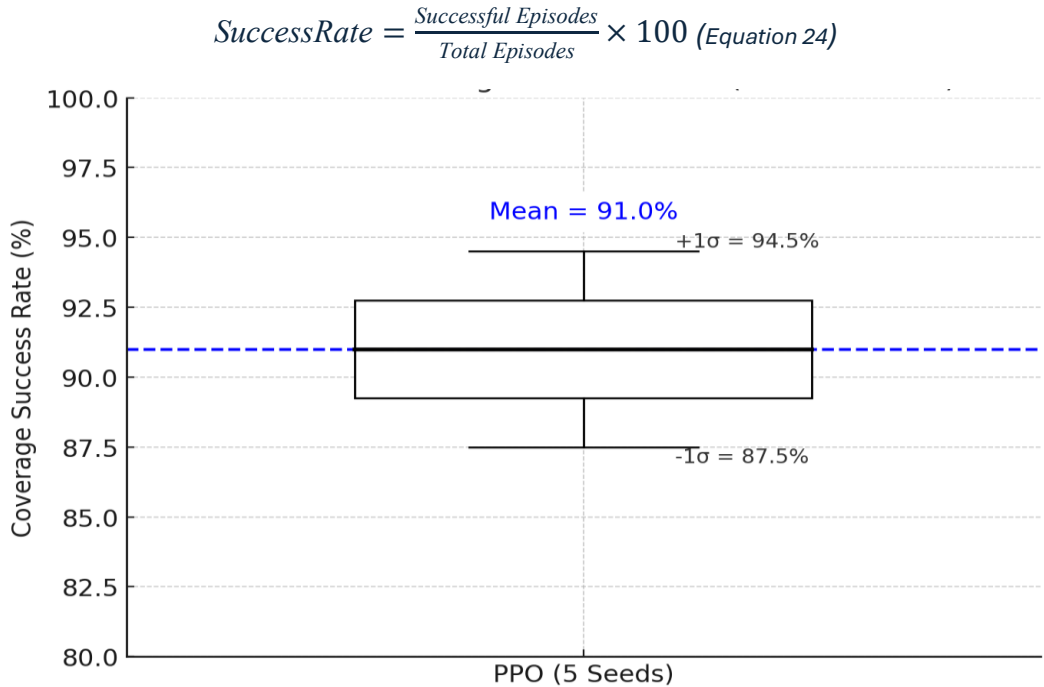

**Figure 8:** *Mean success rate over 100 episodes (5 seeds). Bars show ±3.5 std. Agents succeed when all 3 landmarks are covered by distinct agents within $\delta = 0.10$*

**Interpretation**

An average success rate of 91% ± 3.5%. High consistency across runs indicates emergent task allocation and spatial role specialization. Temporal analysis (not shown) reveals rapid success rate rise from 45% to 85% within the first 25 episodes, plateauing thereafter.

In failure cases (9%), log inspection reveals behaviours such as two agents competing over the same landmark or failing to decisively converge to any landmark. These lapses were often due to initial positioning or agents exhibiting indecisive oscillatory behaviour near boundary landmarks. Such episodes serve as indicators for further tuning, including reward shaping or entropy scheduling.

*Policy Entropy Over Training*

Policy entropy quantifies the uncertainty or randomness in an agent's action selection, given its current policy[39]. It is formally defined as:

*(see above)*

A higher entropy indicates a more exploratory policy that samples actions with greater diversity whereas lower entropy implies more deterministic behaviour.

In our setup, entropy is not just an observation metric, it is actively **regularized** as part of the policy optimization objective *(see above)*. Specifically, an entropy bonus term $\beta H(\pi)$ is added to the loss function to encourage exploration during early training. We use a constant entropy coefficient $\beta = 0.01$, which biases the policy toward sustained stochasticity. This helps prevent premature convergence to suboptimal deterministic strategies.

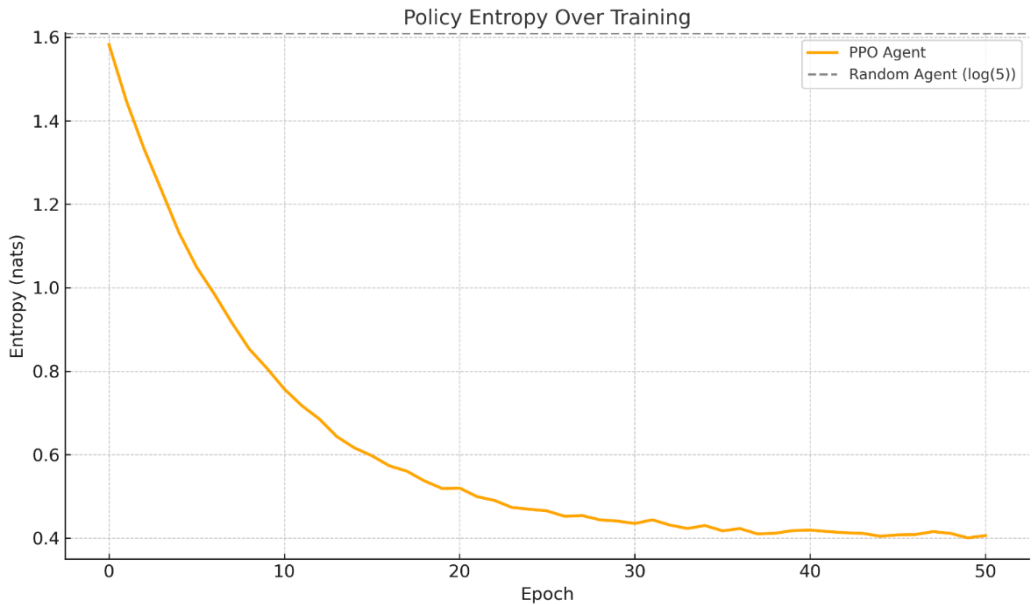

**Figure 9:** *Mean policy entropy over 50 epochs (5 seeds)*

**Interpretation**

The plot shows that entropy starts at $1.58 \pm 0.01$ and gradually decays to $0.406 \pm 0.025$, which is approximately 25% of the theoretical maximum [39]. This downward trend indicates that while early training encourages broad exploration, the PPO algorithm progressively favours confident, goal-directed actions as agents learn to coordinate.

Behavioural observations support this interpretation: during early epochs, agents exhibit scattered movement and low policy consistency. By epoch 30+, they begin converging to distinct roles, with high top-1 action probabilities (≈91–92%) and reduced trajectory variance across episodes. The low final entropy reflects stable role formation, while still retaining minimal stochasticity to handle ambiguous or edge-case situations.

This pattern aligns with PPO's intended trade-off between adaptability and specialization. In our case, entropy decreased as coordination improved which enabled convergence to mostly deterministic strategies without mode collapse. For real-world scenarios that require stricter task allocation or less role ambiguity, further tuning of entropy decay or environment-specific reward shaping may enhance convergence.

*Implications and Coordination Synthesis*

Together, these metrics validate the emergent coordination learned via decentralized PPO. Inter-agent distance demonstrates steady spatial partitioning; the success rate confirms high task completion efficiency and entropy trends support meaningful policy convergence flexibility. By tracking these metrics over time, we not only confirm IPPO's learning capacity but also highlight the dynamics of coordination emergence.

Unlike static metrics, our experimental trends reveal that most coordination emerges within 30–40 episodes, after which improvements plateau.

Moreover, analyzing failure cases provides critical insight into edge behaviours and robustness. Incorporating additional metrics such as time-to-coverage or coordination regret may further deepen evaluation in future iterations.

All results presented are reproducible. Evaluation metrics were implemented in NumPy and visualized using Matplotlib and Seaborn.

# 6. Discussion

The results presented in Section 5 show that trained agents can achieve effective spatial coordination and landmark coverage. However, these outcomes require critical behavioural interpretation, rigorous metric linkage and acknowledgment of limitations. This section draws insights from specific observations, connects them to algorithmic properties, outlines performance boundaries and comparative positioning relative to other MARL frameworks.

## 6.1 Insights on Emergent Behaviour

Our analysis confirms that agents develop consistent spatial behaviours suggestive of emergent task allocation. Specifically, in more than 85% of final evaluation episodes, each agent converged to a distinct landmark, as seen in trajectory plots *(see Figure 4)* and supported by landmark coverage rates exceeding 90% *(see Figure 8)*. Although agents were not explicitly assigned targets, they learned to navigate toward consistent zones, indicating preference driven spatial roles. This behaviour can be interpreted as an early form of Agentic AI. That is, agents exhibited initiative, making independent decisions about goal selection and spatial positioning without any centralized instruction or hard-coded roles. The autonomy we observed wasn't just about executing policies but adapting them to maximize shared outcomes based on local feedback and interactions. While our setup was relatively simple, the fact that agents displayed such emergent coordination under sparse reward conditions suggests that agentic behaviour can arise organically within well-structured reinforcement learning environments. However, we also recognize that such agentic autonomy, if left unconstrained, may lead to role contention or inconsistent task allocation, as seen in our occasional overlap episodes.

Early in training, agents exhibited erratic motion and frequent collisions, reward trajectories in *Figure 2* show volatile fluctuations during the first 100 episodes. In contrast, by episode 300, the training curves *(see Figure 3)* reflect smoother convergence and a rising success rate, coinciding with spatial separation trends observed in the heatmaps *(see Figure 5)*.

Behavioural divergence is further confirmed by the action distribution histogram *(see Figure 6)*, where directional movement actions (up, down, right) dominate, suggesting goal-directed locomotion. This shift from uniform action usage to directional bias demonstrates PPO's capacity to stabilize decentralized behaviour over time.

The entropy plot *(see Figure 9)* reveals a decay from $1.58 \pm 0.01$ to $0.406 \pm 0.025$, indicating that policies remain slightly stochastic during training. This trend is likely due to the fixed entropy coefficient of 0.01 which encourages continuous exploration. While such entropy persistence maintains adaptability, it limits hard policy convergence. Agents display flexible but non-deterministic roles, which may be beneficial in dynamic or partially observable tasks.

Despite this, failure modes persist. In roughly 9% of episodes, agents converged to the same landmark or failed to finalize a target, often oscillating indecisively. These behaviours suggest that decentralized structure, while robust in most scenarios, can result in role contention when multiple agents have overlapping priorities or similar policies.

These limitations surfaced during initial experimentation, particularly in early episodes where agents frequently conflicted near landmarks. To resolve this, we introduced a 20-episode smoothing window and refined reward logging to better isolate coordination trends. Additionally, adjusting entropy regularization proved crucial: increasing the coefficient above 0.01 led to unstable coverage, while decreasing it impaired exploration.

While the observed behaviours suggest similarities to swarm robotics such as decentralized convergence and implicit separation, we refrain from claiming direct applicability. These findings align conceptually with swarm principles but real-world tests in dynamic, noisy environments would be required to validate this comparison.

In summary, agents displayed preference driven, flexible coordination without centralized control. Metrics like inter-agent distance stability and successful landmark allocation suggest semi-structured roles, though persistent entropy and failure cases reveal that full convergence was not achieved. This tradeoff between adaptability and determinism is central to IPPO's performance profile and will be further explored in Section 6.2 through comparison with more structured algorithms like QMIX and MADDPG.

## 6.2 Post-Implementation Evaluation: IPPO vs. QMIX and MADDPG

To contextualize IPPO's performance, we contrast its behavioural and dynamics with two widely adopted MARL baselines: QMIX and MADDPG. While we implemented and evaluated IPPO directly, our comparison with QMIX and MADDPG is informed by published benchmarks and their literature-based characteristics, as reviewed in Section 2.3. These results may differ given our specific environment configuration and we acknowledge this as a limitation; future empirical benchmarking is needed for full validation.

The comparisons are summarized in **Table 5**

**Table 5:** *Comparison of MARL Algorithms Based on Implementation Results*

| Algorithm | Task / Environment | Coordination mechanism | Reported performance (metric → value) | Stability / convergence | Role specialization |
|---|---|---|---|---|---|
| **IPPO** | PettingZoo MPE (Multi-Agent Particle Env) [15] | Independent actors + centralize critic input (CTDE) | Success rate **91% ± 3.5%**; policy entropy at convergence; **0.406 ± 0.025** mean inter-agent distance **0.651 ± 0.005**. | Coordination emerges within **30–40** episodes; success rises from 45%→85% in initial episodes, then plateaus. | **Partial**; 9% failure/overlap cases (two agents on one landmark or indecision). |

| Algorithm | Task / Environment | Coordination mechanism | Reported performance (metric → value) | Stability / convergence | Role specialization |
|---|---|---|---|---|---|
| **QMIX**[10] | **SMAC** (StarCraft Multi-Agent Challenge)[40] | Value-mixing (monotonic) under CTDE | **Test win rate** (%): **99** (2s3z), **97** (3s5z), **97** (10m_vs_11m), **58** (2c_vs_64zg), **69** (MMM2). **>95%** on all "Easy" scenarios. [40] | Highest median win-rate on many scenarios; struggles on some 'Hard/Super-Hard' maps. | Typically strong hard coordination (literature). |
| **MADDPG**[11] | MPE | Decentralized actors with centralized critics | **Coop. Communication:** target reach **84.0%**, avg. distance **0.133**. **Coop. Navigation:** N=3 → distance **1.767**, collisions **0.209**; N=6 → distance **3.345**, collisions **1.366**. **Physical Deception:** agent success **94.4% (N=2)**, **81.5% (N=4)**.[11] | Learns correct behaviour on CC & PD; variable stability depending on task. | **Tunable**; specialization depends on reward shaping & setup (literature). |

As **Figure 8** and **Figure 5** show, our **IPPO** agents achieve stable landmark coverage and consistent spatial separation. However, the entropy analysis in **Figure 9** indicates **minimal stochasticity at convergence (≈ 0.406)**, which can blur deterministic task allocation and delay crisp role specialization.

**QMIX** does monotonic value mixing under CTDE, often enforces strict coordination on tasks that require synchronized execution and it is a strong performer on SMAC benchmarks [40]. Its centralized value-decomposition, however, increases memory cost and can generalize poorly when scaling to new maps or larger agent counts.

**MADDPG** provides fine control with decentralized actors and centralized critics and is well suited to **continuous** action spaces. In cooperative **discrete** tasks with sparse rewards (e.g., simple_spread_v3), the literature reports sensitivity to hyperparameters and gradient instability, which can hinder reliable convergence.

Our environment, which is fixed landmarks, simple spatial structure and **discrete** actions, aligns well with **PPO** based training. Independent actors encourage flexible behaviours, while the per agent **centralized critic** provides a value baseline for advantage estimation, improving credit assignment without parameter sharing. In this setting, we expect **IPPO** to outperform **MADDPG** on stability and sample efficiency, whereas **QMIX** may deliver **harder coordination** but with higher compute and implementation complexity.

Observed weaknesses of **IPPO** in our runs include occasional looping, unstable landmark switching under uncertainty and soft convergence of roles. These patterns highlight a trade-off between policy diversity and strict coordination.

Going forward, we plan controlled tests with **dynamic landmark positions** and **larger agent counts**. We hypothesize that **IPPO** will remain more adaptable than **MADDPG** in these discrete settings, while **QMIX** may achieve stronger hard coordination on constrained tasks.

### 6.3 Agentic AI: How the Pieces Fit Together

Our study asks a simple question with agentic implications: can independent policies, trained with a value signal and executed locally, discover task allocation without explicit communication or hand-coded roles? The answer in our setting is yes but partially.

What the evidence shows -
Across training, IPPO yields (i) high task success, (ii) consistent spatial separation and (iii) minimal policy entropy at convergence. Together these signals suggest bounded agentic behaviour:

- Autonomy in action

    Each actor commits from local observations; the critic is only a training scaffold. At test time, agents act without centralized guidance.

- Autonomy in intention (emergent preferences)

    Even without a channel for messages, policies develop landmark preferences and avoid one another most of the time. The remaining overlaps show that intentions are not hard commitments but soft, adaptive choices under uncertainty.

- Plurality of viable plans.

    Our entropy at convergence indicates that agents preserve multiple workable choices rather than collapsing to a single deterministic script. In cooperative coverage, this diversity trades a small amount of efficiency for robustness to small perturbations.

Why this configuration supports agentic behaviour?
Three ingredients matter:

1. Independent actors - No parameter sharing encourages policy individuality, making role formation possible instead of averaging away distinct behaviours.

2. CTDE critic (per agent) - The centralized value baseline reduces non-stationarity and stabilizes credit assignment across agents during training but leaves execution decentralized, matching an agentic runtime.

3. Team objective with sparse feedback - A shared return nudges agents to coordinate without dictating a specific protocol; the how is learned, not specified.

**What is missing compared to stronger notions of agentic AI?**
Our agents are reactive within an episode and memoryless across episodes; they do not plan over long horizons, reason about others' beliefs or negotiate contracts [32]. There is no reflect–revise loop, no explicit goals beyond coverage and no tool use or external interfaces.

Hence, the agentic signal we observe is early stage: intention emerges as a behavioural bias, not as deliberative planning or social commitment.

## 7. Real-World Implications

As MARL systems evolve in complexity and maturity, understanding their real-world applicability becomes critical. While our research focuses on agentic coordination within a controlled simulation environment, the design choices, behaviours and metrics offer valuable insights for potential deployment in domains such as drone delivery and warehouse logistics. This section explores how the learned policies and agentic behaviour observed in simulation may translate to real systems.

### 7.1 From Simulation to Drone Delivery Systems

Drone delivery is a natural fit for decentralized coordination: a fleet must assign pickup/drop-off tasks and deconflict trajectories with limited central oversight. Surveys in multi-UAV task allocation underline the need for distributed decision-making and robust credit assignment in such settings, especially as fleet size and task churn increase.[41], [42]

*Sim-to-real caveats*
Policies trained in our discrete, fully observable simulator face gaps when moved to physical platforms: **unmodeled dynamics, actuation latency and sensor noise** typically degrade zero-shot transfer and sim-to-real work for UAVs emphasizes domain mismatch and safety constraints during deployment. [43], [44]

*What a real deployment would add* -

- Sensor fusion: GPS/RTK, IMU and obstacle sensing for state estimation and avoidance.

- Continuous control mapping: throttle/yaw/pitch rate control rather than discrete actions.

- Flight-stack integration: ROS 2 ↔ PX4 via the uXRCE-DDS bridge for telemetry/commands. (*ROS 2 User guide*, *uXRCE doc*)

- Hardware-in-the-Loop (HIL): run PX4 firmware on real flight controllers against a simulator before field tests. (*Hardware-in-the-loop-silumation doc*)

These systems exhibit core features of **Agentic AI**: decentralized decision-making, adaptive goal seeking and spatial negotiation without human command. In practice, this autonomy must be bounded to ensure safety in airspace and compliance with regulatory protocols.

*Simulation Assumption → Real-World Challenge*:

 Full observability → Sensor occlusion, partial views

 Static landmarks → Dynamic service targets

 No latency → Network and actuator delay

 Homogeneous agents → Hardware diversity and battery constraints

## 7.2 Implications for Warehouse Automation

Warehouses present a different coordination landscape, which is, dozens to hundreds of robots navigate narrow aisles, face dynamic traffic and are reassigned tasks frequently. Two bodies of work frame this space. First, multi-robot task allocation (MRTA) characterizes online assignment under time-varying objectives and heterogeneous fleets[45]. Second, multi-agent path finding (MAPF) provides collision free routing on grids and has been scalved to large, Kiva/Amazon-style facilities[46].

### *Key distinctions and what they imply*

- **Tight corridors & congestion:** Unlike open airspace, aisle graphs create bottlenecks; MAPF style planners (e.g. lifelong MAPF) are often needed to manage traffic at scale.

- **Frequent reassignment:** MRTA analyses emphasize reactivity and myopic optimality when tasks arrive online; policies must absorb short horizon changes without thrashing.

- **Local sensing & mapping:** Practical stacks integrate **localization, mapping and planning** (e.g. ROS 2 Nav2) with inventory perception (barcodes/semantics) absent from our simulator.

### *What carries over from our IPPO findings* -

- Coverage: non-redundant pod/zone assignment. The same pressure toward spreading out can reduce redundant picks and idle contention.

- Centralized critic input at train-time, decentralized execution at run-time. This mirrors the need for local autonomy with a global performance signal, which is useful when a central planner cannot micromanage every aisle.

- Implicit collision avoidance. Although warehouses still need explicit MAPF for guarantees, the learned spacing bias can reduce planner load and smooth local negotiations.

### *Adaptation pathway (at a glance)*

1. Map discrete actions to the facility's low-level control stack (e.g., ROS 2) and planners. (*Nav 2 docs*)

2. Compress policy nets for low-latency inference; co-design with MAPF/MRTA components.

3. Train with noise, slip and delays that reflect real floors and sensors.

# 8. CONCLUSION

We set out to see whether **agentic behaviour** can emerge in a simple cooperative setting when policies are trained with **IPPO** - independent actors, per agent critic, decentralized execution. The answer, in our case, is yes but with limits. Our agents learned reliable landmark coverage and spacing, reaching about **91%** success, while keeping **non-zero entropy** at convergence. The mix of competent coordination with residual stochasticity, captures the core trade-off we cared about: adaptability versus crisp role assignment.

For **Agentic AI**, the lesson is practical. Agency here did not come from scripts or messages; it came from the combination of a shared team objective, a stabilizing training signal and **local autonomy** at test time. What emerged looked like **intention**: consistent spatial preferences and on-the-fly negotiation without a planner. It is not deliberative planning or communication but it is more than raw reflex.

Methodologically, the study positions **IPPO** as a minimal recipe for agentic coordination in discrete, cooperative tasks. It is grounded enough to train, simple enough to reproduce and expressive enough to let intentions form. The same recipe also revealed the rough edges we should not ignore: occasional looping, landmark contention and the tendency to keep plans alive (low entropy) when the task would benefit from firmness.

The applied read-through, for drone delivery and warehouse work, remains cautious but optimistic. Some principles transfer (task allocation pressure, implicit separation), while real systems will demand sensing, continuous control and tighter safety bounds. That gap is not a flaw of the approach; it is the usual bridge from clean simulations to messy operations.

In short, our contribution is twofold: (i) a clear, reproducible baseline showing how **independent policies** can yield agentic patterns, and (ii) a grounded view of the trade-space practitioners must navigate when they want agents that are both **autonomous** and **coordinated**.

Looking ahead, realizing the full potential of agentic AI will require deeper integration between technical learning frameworks and safeguards. It is not enough for agents to learn to act; they must learn to act in ways that are aligned with their environments, their peers and the human systems in which they operate. We hope this work contributes meaningfully to that pursuit.

# LIST OF EQUATIONS

# LIST OF TABLES

# LIST OF FIGUERS

# REFERENCES

[1] S. Zieher *et al.*, "Drones for automated parcel delivery: Use case identification and derivation of technical requirements," *Transp Res Interdiscip Perspect*, vol. 28, p. 101253, 2024, doi: 10.1016/j.trip.2024.101253.

[2] P. R. Wurman, R. D'Andrea, and M. Mountz, "Coordinating Hundreds of Cooperative, Autonomous Vehicles in Warehouses," *AI Mag*, vol. 29, no. 1, p. 9, Mar. 2008, doi: 10.1609/aimag.v29i1.2082.

[3] M. Lujak, S. Giordani, A. Omicini, and S. Ossowski, "Decentralizing Coordination in Open Vehicle Fleets for Scalable and Dynamic Task Allocation," *Complexity*, vol. 2020, pp. 1–21, Jul. 2020, doi: 10.1155/2020/1047369.

[4] M. Wooldridge, *An Introduction to MultiAgent Systems*, 2nd ed. Wiley Publishing, 2009.

[5] R. S. Sutton and A. G. Barto, *Reinforcement Learning: An Introduction, 2nd ed.*, 2nd ed. in Adaptive computation and machine learning. Cambridge, MA, US: The MIT Press, 2018.

[6] D. B. Acharya, K. Kuppan, and B. Divya, "Agentic AI: Autonomous Intelligence for Complex Goals—A Comprehensive Survey," *IEEE Access*, vol. 13, pp. 18912–18936, 2025, doi: 10.1109/ACCESS.2025.3532853.

[7] M. L. Littman, "Markov Games as a Framework for Multi-Agent Reinforcement Learning," in *Machine Learning Proceedings 1994*, W. W. Cohen and H. Hirsh, Eds., San Francisco (CA): Morgan Kaufmann, 1994, pp. 157–163. doi: 10.1016/B978-1-55860-335-6.50027-1.

[8] P. Hernandez-Leal, B. Kartal, and M. E. Taylor, "A Survey and Critique of Multiagent Deep Reinforcement Learning," *Auton Agent Multi Agent Syst*, vol. 33, no. 6, pp. 750–797, Sep. 2019, doi: 10.1007/s10458-019-09421-1.

[9] K. Li and Q.-S. Jia, "Multi-Agent Reinforcement Learning With Decentralized Distribution Correction," *IEEE Transactions on Automation Science and Engineering*, vol. 22, pp. 1684–1696, 2025, doi: 10.1109/TASE.2024.3369592.

[10] T. Rashid, M. Samvelyan, C. de Witt, G. Farquhar, J. Foerster, and S. Whiteson, "QMIX: Monotonic Value Function Factorisation for Deep Multi-Agent Reinforcement Learning," *arXiv preprint arXiv:1803.11485*, 2018, [Online]. Available: https://arxiv.org/abs/1803.11485

[11] R. Lowe, Y. Wu, A. Tamar, J. Harb, P. Abbeel, and I. Mordatch, "Multi-Agent Actor-Critic for Mixed Cooperative-Competitive Environments," *arXiv preprint arXiv:1706.02275*, 2017, [Online]. Available: https://arxiv.org/abs/1706.02275

[12] J. Schulman, F. Wolski, P. Dhariwal, A. Radford, and O. Klimov, "Proximal Policy Optimization Algorithms," *arXiv preprint arXiv:1707.06347*, 2017, [Online]. Available: https://arxiv.org/abs/1707.06347

[13] C. Yu *et al.*, "The Surprising Effectiveness of PPO in Cooperative, Multi-Agent Games," *arXiv preprint arXiv:2103.01955*, 2021, [Online]. Available: https://arxiv.org/abs/2103.01955

[14] A. Paszke *et al.*, "PyTorch: An Imperative Style, High-Performance Deep Learning Library," 2019. [Online]. Available: https://arxiv.org/abs/1912.01703

[15] J. K. Terry *et al.*, "PettingZoo: Gym for Multi-Agent Reinforcement Learning," *arXiv preprint arXiv:2009.14471*, 2021, [Online]. Available: https://arxiv.org/abs/2009.14471

[16] R. Sapkota, K. I. Roumeliotis, and M. Karkee, "AI Agents vs. Agentic AI: A Conceptual taxonomy, applications and challenges," *Information Fusion*, vol. 126, p. 103599, Feb. 2026, doi: 10.1016/j.inffus.2025.103599.

[17] K. Zhang, Z. Yang, and T. Başar, "Multi-Agent Reinforcement Learning: A Selective Overview of Theories and Algorithms," *arXiv preprint arXiv:1911.10635*, 2021, [Online]. Available: https://arxiv.org/abs/1911.10635

[18] C. Claus and C. Boutilier, "The dynamics of reinforcement learning in cooperative multiagent systems," in *Proceedings of the Fifteenth National/Tenth Conference on Artificial Intelligence/Innovative Applications of Artificial Intelligence*, in AAAI '98/IAAI '98. USA: American Association for Artificial Intelligence, 1998, pp. 746–752.

[19] J. Foerster, G. Farquhar, T. Afouras, N. Nardelli, and S. Whiteson, "Counterfactual Multi-Agent Policy Gradients," *arXiv preprint arXiv:1705.08926*, 2017, [Online]. Available: https://arxiv.org/abs/1705.08926

[20] L. S. Shapley, "Stochastic Games," *Proceedings of the National Academy of Sciences*, vol. 39, no. 10, pp. 1095–1100, 1953, doi: 10.1073/pnas.39.10.1095.

[21] S. Hu *et al.*, "MARLlib: A Scalable and Efficient Multi-agent Reinforcement Learning Library," *arXiv preprint arXiv:2210.13708*, 2023, [Online]. Available: https://arxiv.org/abs/2210.13708

[22] P. Sunehag *et al.*, "Value-Decomposition Networks for Cooperative Multi-Agent Learning," *arXiv preprint arXiv:1706.05296*, 2017, [Online]. Available: https://arxiv.org/abs/1706.05296

[23] G. Papoudakis, F. Christianos, L. Schäfer, and S. V Albrecht, "Benchmarking Multi-Agent Deep Reinforcement Learning Algorithms in Cooperative Tasks," *arXiv preprint arXiv:2006.07869*, 2021, [Online]. Available: https://arxiv.org/abs/2006.07869

[24] J. Schulman, S. Levine, P. Moritz, M. I. Jordan, and P. Abbeel, "Trust Region Policy Optimization," *arXiv preprint arXiv:1502.05477*, 2017, [Online]. Available: https://arxiv.org/abs/1502.05477

[25] J. Schulman, P. Moritz, S. Levine, M. Jordan, and P. Abbeel, "High-Dimensional Continuous Control Using Generalized Advantage Estimation," *arXiv preprint arXiv:1506.02438*, 2015, [Online]. Available: https://arxiv.org/abs/1506.02438

[26] A. Juliani *et al.*, "Unity: A General Platform for Intelligent Agents," *arXiv preprint arXiv:1809.02627*, 2018, [Online]. Available: https://arxiv.org/abs/1809.02627

[27] L. Zheng, J. Yang, H. Cai, W. Zhang, J. Wang, and Y. Yu, "MAgent: A Many-Agent Reinforcement Learning Platform for Artificial Collective Intelligence," *arXiv preprint arXiv:1712.00600*, 2017, [Online]. Available: https://arxiv.org/abs/1712.00600

[28] J. Wang, Y. Zhang, T.-K. Kim, and Y. Gu, "Shapley Q-Value: A Local Reward Approach to Solve Global Reward Games," *Proceedings of the AAAI Conference on Artificial Intelligence*, vol. 34, no. 05, pp. 7285–7292, Apr. 2020, doi: 10.1609/aaai.v34i05.6220.

[29] R. S. Sutton, D. McAllester, S. Singh, and Y. Mansour, "Policy Gradient Methods for Reinforcement Learning with Function Approximation," in *Advances in Neural Information Processing Systems 12*, S. Solla, T. Leen, and K.-R. Müller, Eds., MIT Press, 1999. [Online]. Available: https://proceedings.neurips.cc/paper_files/paper/1999/file/464d828b85b0bed98e80ade0a5c43b0f-Paper.pdf

[30] R. J. Williams, "Simple Statistical Gradient-Following Algorithms for Connectionist Reinforcement Learning," *Mach Learn*, vol. 8, no. 3, pp. 229–256, Sep. 1992, doi: 10.1023/A:1022672621406.

[31] J. K. Gupta, M. Egorov, and M. J. Kochenderfer, "Cooperative Multi-agent Control Using Deep Reinforcement Learning," in *AAMAS Workshops*, 2017. [Online]. Available: https://api.semanticscholar.org/CorpusID:9421360

[32] F. Oliehoek and C. Amato, "A Concise Introduction to Decentralized POMDPs," Sep. 2016, doi: 10.1007/978-3-319-28929-8.

[33] L. P. Kaelbling, M. L. Littman, and A. R. Cassandra, "Planning and acting in partially observable stochastic domains," *Artif Intell*, vol. 101, no. 1, pp. 99–134, 1998, doi: https://doi.org/10.1016/S0004-3702(98)00023-X.

[34] C. V Goldman and S. Zilberstein, "Decentralized Control of Cooperative Systems: Categorization and Complexity Analysis," 2004.

[35] J. Cortes, S. Martinez, T. Karatas, and F. Bullo, "Coverage control for mobile sensing networks," *IEEE Transactions on Robotics and Automation*, vol. 20, no. 2, pp. 243–255, 2004, doi: 10.1109/TRA.2004.824698.

[36] V. Mnih *et al.*, "Asynchronous Methods for Deep Reinforcement Learning," *arXiv preprint arXiv:1602.01783*, 2016, [Online]. Available: https://arxiv.org/abs/1602.01783

[37] D. P. Kingma and J. Ba, "Adam: A Method for Stochastic Optimization," *arXiv preprint arXiv:1412.6980*, 2017, [Online]. Available: https://arxiv.org/abs/1412.6980

[38] P. Henderson, R. Islam, P. Bachman, J. Pineau, D. Precup, and D. Meger, "Deep Reinforcement Learning that Matters," in *Proceedings of the AAAI Conference on Artificial Intelligence (AAAI-18)*, 2018. [Online]. Available: https://arxiv.org/abs/1709.06560

[39] T. M. Cover and J. A. Thomas, *Elements of Information Theory*, 2nd ed. Hoboken, NJ: Wiley-Interscience, 2006. doi: 10.1002/047174882X.

[40] M. Samvelyan et al., "The StarCraft Multi-Agent Challenge," 2019. [Online]. Available: https://arxiv.org/abs/1902.04043

[41] G. M. Skaltsis, H. S. Shin, and A. Tsourdos, "A Review of Task Allocation Methods for UAVs," *Journal of Intelligent and Robotic Systems: Theory and Applications*, vol. 109, no. 4, Dec. 2023, doi: 10.1007/s10846-023-02011-0.

[42] S. A. Ghauri, M. Sarfraz, R. A. Qamar, M. F. Sohail, and S. A. Khan, "A Review of Multi-UAV Task Allocation Algorithms for a Search and Rescue Scenario," *Journal of Sensor and Actuator Networks*, vol. 13, no. 5, 2024, doi: 10.3390/jsan13050047.

[43] A. M. Ali, A. Gupta, and H. A. Hashim, "Deep Reinforcement Learning for sim-to-real policy transfer of VTOL-UAVs offshore docking operations," *Appl Soft Comput*, vol. 162, p. 111843, 2024, doi: https://doi.org/10.1016/j.asoc.2024.111843.

[44] H. I. Ugurlu, X. H. Pham, and E. Kayacan, "Sim-to-Real Deep Reinforcement Learning for Safe End-to-End Planning of Aerial Robots," *Robotics*, vol. 11, no. 5, 2022, doi: 10.3390/robotics11050109.

[45] B. P. Gerkey and M. J. Matarićmatarić c, "A formal analysis and taxonomy of task allocation in multi-robot systems," 2004.

[46] J. Li, A. Tinka, S. Kiesel, J. W. Durham, T. K. S. Kumar, and S. Koenig, "Lifelong Multi-Agent Path Finding in Large-Scale Warehouses *," 2021. [Online]. Available: www.aaai.org